\pdfoutput=1

\documentclass[11pt]{article}

\usepackage[]{acl}

\usepackage{times}
\usepackage{latexsym}
\usepackage{amsfonts}
\usepackage{booktabs}
\usepackage{multirow}
\usepackage{makecell}
\usepackage{amsmath}
\usepackage{subfigure}
\usepackage{booktabs}
\usepackage{adjustbox}

\usepackage[T1]{fontenc}

\usepackage[utf8]{inputenc}

\usepackage{microtype}

\usepackage{inconsolata}

\usepackage{graphicx}

\title{Locate-then-Merge: Neuron-Level Parameter Fusion for \\ 
Mitigating Catastrophic Forgetting in Multimodal LLMs}

\author{Zeping Yu \quad Sophia Ananiadou\\
  Department of Computer Science, National Centre for Text Mining \\
  The University of Manchester  \\
  \texttt{\{zeping.yu@postgrad. sophia.ananiadou@\}manchester.ac.uk}}

\begin{document}
\maketitle
\begin{abstract}
Although multimodal large language models (MLLMs) have achieved impressive performance, the multimodal instruction tuning stage often causes catastrophic forgetting of the base LLM’s language ability, even in strong models like Llama3. To address this, we propose Locate-then-Merge, a training-free parameter fusion framework that first locates important parameters and then selectively merges them. We further introduce Neuron-Fusion, a neuron-level strategy that preserves the influence of neurons with large parameter shifts—neurons likely responsible for newly acquired visual capabilities—while attenuating the influence of neurons with smaller changes that likely encode general-purpose language skills. This design enables better retention of visual adaptation while mitigating language degradation. Experiments on 13 benchmarks across both language and visual tasks show that Neuron-Fusion consistently outperforms existing model merging methods. Further analysis reveals that our method effectively reduces context hallucination in generation.
\end{abstract}

\section{Introduction}
Multimodal large language models (MLLMs) \citep{liu2023visual,team2023gemini,chen2024internvl,wu2024deepseek,hurst2024gpt,bai2025qwen2} have advanced rapidly by adapting pretrained large language models (LLMs) \cite{brown2020language,ouyang2022training,yang2024qwen2,grattafiori2024llama} through multimodal instruction tuning. Among various modalities, vision has received the most attention and become the primary focus for enhancing multimodal LLMs \cite{liang2024survey,li2025benchmark}. By introducing vision-language connectors and training with image-text pairs, MLLMs have demonstrated impressive performance on vision-language tasks such as visual question answering \citep{antol2015vqa} and visual reasoning \cite{hudson2019gqa}.

However, recent studies \citep{ratzlaff2024training,zhang2024wings} find that although visual instruction tuning can obtain visual capabilities, it often severely degrades the original general language abilities of the base LLMs, a phenomenon known as catastrophic forgetting \citep{goodfellow2013empirical,kirkpatrick2017overcoming,kemker2018measuring}. Particularly on complex language understanding and reasoning benchmarks such as ARC-Challenge \citep{clark2018think} and GSM8K \citep{cobbe2021training}, finetuned MLLMs perform significantly worse than their original LLMs. Alarmingly, even strong open-source models like Llama3 suffer from this degradation, limiting the generalization and practical deployment of MLLMs across diverse and complicated language tasks. 

\begin{figure}[thb]
  \centering
  \includegraphics[width=0.98\columnwidth]{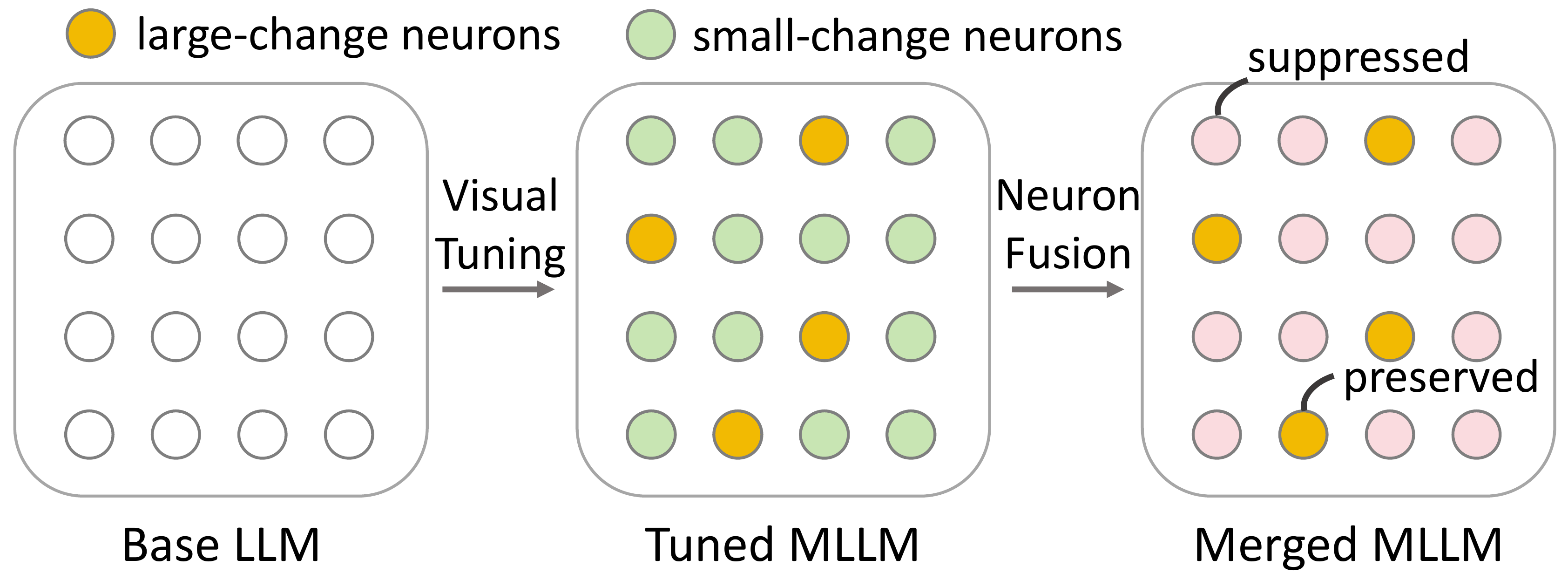}
  \caption{Neuron-Fusion in MLLMs. After visual tuning, some neurons exhibit larger changes than others. Neuron-Fusion selectively preserves neurons with significant parameter changes while suppressing those with smaller changes. This targeted fusion enables the model to retain newly acquired visual capabilities while minimally affecting its general language abilities.}
\vspace{-10pt}
\end{figure}

Although catastrophic forgetting in MLLMs poses a significant challenge, systematic studies addressing this issue remain limited. Recent years have witnessed the emergence of several novel model merging techniques \cite{wortsman2022model,ilharco2022editing,yadav2023ties,yu2024language}, demonstrating the potential of parameter fusion to alleviate catastrophic forgetting. However, as an emerging research area, model merging still lacks a systematic framework to guide the design and evaluation of effective methods. Moreover, existing methods were primarily developed for single-modal LLMs, and their application to MLLMs remains largely unexplored. It is unclear whether these methods can effectively mitigate catastrophic forgetting in MLLMs. This highlights the urgent need for targeted solutions that can recover language abilities while preserving visual adaptation. 

In this paper, we propose Locate-then-Merge, a general framework for training-free parameter fusion that decouples the process into two stages: locating important parameters and selectively merging them. Within this framework, we further develop Neuron-Fusion, a neuron-level fusion method designed to mitigate catastrophic forgetting, as illustrated in Figure 1. Our approach is motivated by an intuitive hypothesis: neurons with large changes during visual instruction tuning are likely to store newly acquired visual capabilities, whereas widespread but small changes across many neurons can cumulatively cause catastrophic forgetting in language ability. Therefore, Neuron-Fusion preserves the contributions of large-change neurons while suppressing the influence of small-change neurons. This design aims to retain the general language abilities of the base LLM while maintaining its acquired visual skills. Empirical results demonstrate that Neuron-Fusion surpasses five state-of-the-art model merging methods across 13 language and vision benchmarks, evaluated on two leading open-source MLLMs. Furthermore, generation analysis shows that Neuron-Fusion effectively reduces context hallucination, enhancing output quality and controllability.

Our contributions are summarized as follows:

a) We systematically validate that visual instruction tuning induces catastrophic forgetting even in powerful MLLMs, as evidenced by significant performance degradation on multiple language understanding and reasoning benchmarks, indicating that catastrophic forgetting remains a widespread and persistent challenge in MLLMs.

b) We propose Locate-then-Merge, a general parameter fusion framework that unifies existing merging strategies: locating important parameters and merging them. Within this framework, we introduce Neuron-Fusion, a neuron-level selection and fusion method that identifies a small subset of neurons with significant parameter changes as key paths for visual adaptation, which is useful for mitigating catastrophic forgetting in MLLMs.

c) We conduct extensive experiments across 13 benchmarks and two powerful open-source MLLMs to validate our approach. Our method is compared against several state-of-the-art model merging techniques, consistently demonstrating superior performance in mitigating catastrophic forgetting while preserving visual adaptation.

\section{Related Work}
\subsection{Multimodal Large Language Models}
A typical MLLM consists of three components: a modality encoder, a modality connector, and a pre-trained LLM. In this work, we focus on the vision modality. First, a pre-trained vision encoder such as CLIP \cite{radford2021learning} is employed to transform images into visual embeddings. Second, a vision-language connector, such as a lightweight two-layer MLP \cite{liu2023visual} or cross-attention layers \cite{alayrac2022flamingo}, maps these embeddings into the feature space of the LLM. Finally, the pre-trained LLM, such as Llama3 \cite{grattafiori2024llama}, generates output tokens conditioned on both the textual and visual inputs. During the multimodal instruction tuning phase, the parameters of both the connector and the LLM are jointly finetuned to adapt to vision-language tasks.

\subsection{Catastrophic Forgetting and Model Merging}
Catastrophic forgetting \cite{goodfellow2013empirical} refers to the phenomenon in which a machine learning model loses previously acquired knowledge when learning new capabilities.
\citet{luo2023empirical} observe that catastrophic forgetting frequently occurs in LLMs during finetuning, and that larger models tend to forget even more information.
\citet{zhu2024model} further investigate this phenomenon in the context of visual tasks.
In addition, \citet{zhang2024wings} and \citet{ratzlaff2024training} report that visual instruction tuning not only improves multimodal capabilities but also significantly impairs the general language abilities of the underlying LLMs.
\citet{van2024continual} hypothesize that catastrophic forgetting arises because the parameters learned on new tasks deviate substantially from the optimal parameters for previous tasks.

Model merging has emerged as an effective technique for combining the parameters of different models to construct a universal model, without requiring access to the training data \cite{yang2024model}. \citet{alexandrov2024mitigating} demonstrate that model merging can be beneficial for mitigating catastrophic forgetting. Recent works \cite{wortsman2022model, ilharco2022editing, yadav2023ties, davari2024model, yu2024language, deep2024della} have shown that various merging strategies can significantly improve performance across different tasks and models.

\subsection{Neuron-Level Ability Storage in LLMs}
\citet{geva2020transformer} observe that in transformer models, the columns and rows of the two MLPs in Feed-Forward Network (FFN) layers can be interpreted as key and value memories, respectively. Similarly, \citet{yu2023neuron} find that the attention value-output matrices, which are implemented as two MLPs, can also be understood in terms of neuron representations. \citet{dai2021knowledge} show that factual knowledge is primarily encoded in FFN neurons.
\citet{geva2022transformer} and \citet{lee2024mechanistic} demonstrate that the generation of toxic language can be controlled by editing targeted neurons.
Furthermore, \citet{nikankin2024arithmetic} and \citet{yu2024interpreting} find that arithmetic abilities are localized to a small subset of neurons. \citet{schwettmann2023multimodal} discover the ``multimodal neurons'' in pretrained text-only transformers.

\section{Background and Problem Formulation}
\subsection{Architectures of LLM and MLLM}
The architectures of LLM and MLLM are illustrated in Figure 2. In this study, we focus primarily on the vision modality, although the methods can be similarly applied to other modalities \cite{liang2024survey}. In the original LLM architecture, the base LLM processes textual inputs and generates corresponding textual outputs. In contrast, the MLLM architecture incorporates an additional vision encoder, which transforms images into visual embeddings. These embeddings are then mapped into the feature space of a tuned LLM via a vision-language connector. Finally, the tuned LLM takes both the textual inputs and the mapped visual embeddings as inputs, producing textual outputs. It is important to note that the tuned LLM's parameters are obtained by performing visual instruction tuning on the base LLM using image-text pairs.

\begin{figure}[thb]
  \centering
  \includegraphics[width=0.98\columnwidth]{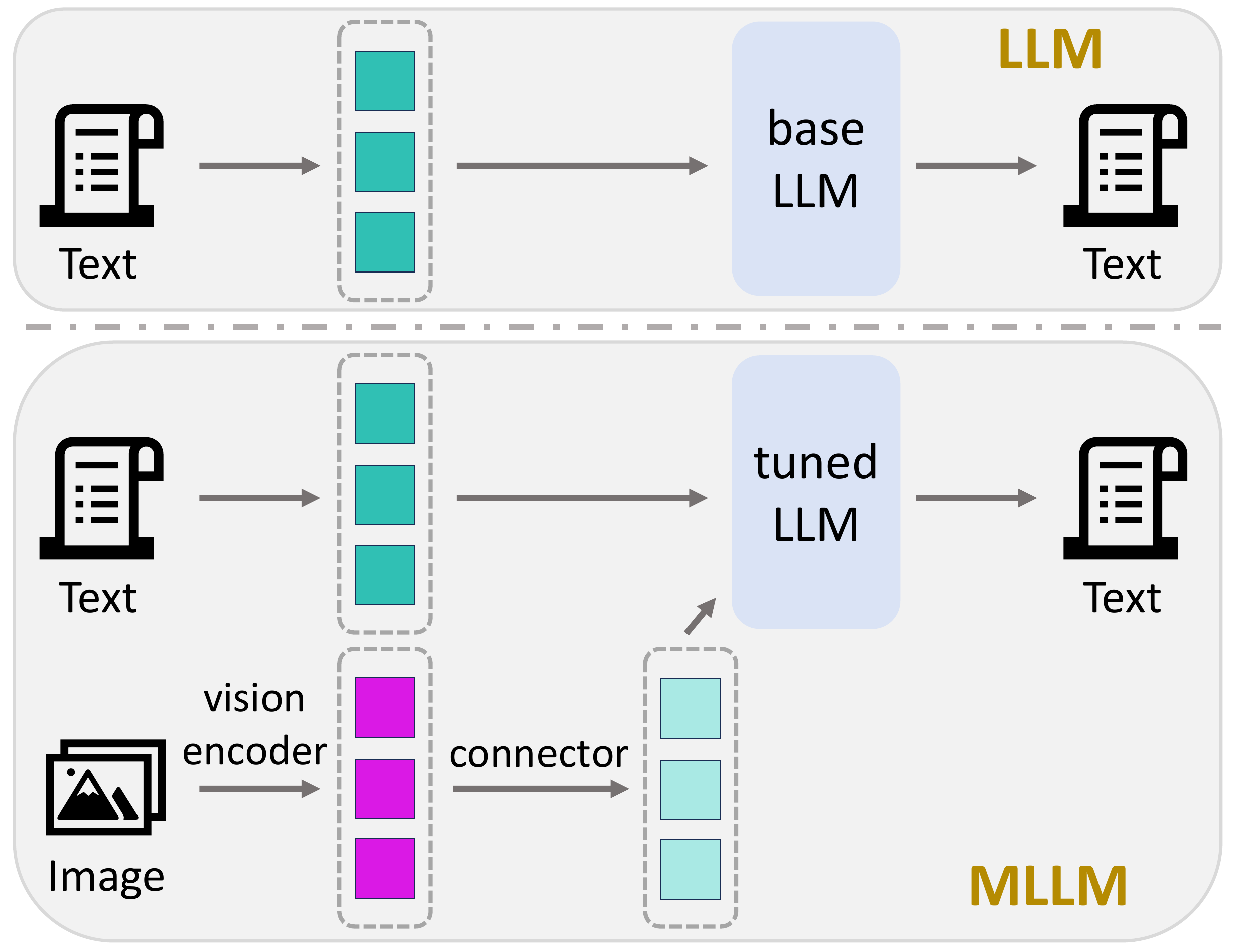}
  \caption{The structures of LLM and MLLM.}
\vspace{-10pt}
\end{figure}

\subsection{Problem Formulation}
Formally, let $LLM_{\text{base}}$ denote the base LLM with parameters $\theta_{\text{base}}$ and language ability $L_{\text{base}}$. 
Similarly, let $LLM_{\text{tuned}}$ denote the tuned LLM in the MLLM, with parameters $\theta_{\text{tuned}}$. We define the parameter delta as $\Delta = \theta_{\text{tuned}} - \theta_{\text{base}}$, representing the changes introduced during visual instruction tuning. After tuning, the MLLM acquires a visual ability $V_{\text{tuned}}$, but its language ability degrades to $L_{\text{tuned}}$ (typically $L_{\text{tuned}} < L_{\text{base}}$) due to catastrophic forgetting. This degradation occurs because some parameters responsible for language capabilities are inadvertently modified during visual instruction tuning. A straightforward way to recover language ability is to replace $LLM_{\text{tuned}}$ with $LLM_{\text{base}}$ in the MLLM, while keeping the vision encoder and the vision-language connector unchanged—a process we refer to as ``hard-merge''. However, hard-merge often severely damages the visual ability, where $V_{\text{base}}$ becomes significantly lower than $V_{\text{tuned}}$.

Our goal is to obtain a merged LLM, denoted as $LLM_{\text{merge}}$, with parameters $\theta_{\text{merge}}$ and language ability $L_{\text{merge}}$. By replacing $LLM_{\text{tuned}}$ with $LLM_{\text{merge}}$ in the MLLM while keeping the vision encoder and connector unchanged, we obtain a visual ability $V_{\text{merge}}$. Ideally, we aim for $L_{\text{merge}}$ to be slightly smaller than, or even comparable to, $L_{\text{base}}$, and for $V_{\text{merge}}$ to be slightly smaller than, or comparable to, $V_{\text{tuned}}$. In this desirable scenario, both the language and visual abilities are preserved.

\section{Methodology}
In order to solve the catastrophic forgetting problem, we aim to utilize model merging methods to obtain $LLM_{\text{merge}}$ based on $LLM_{\text{base}}$ and $LLM_{\text{tuned}}$. We first propose the Locate-then-Merge framework in Section 4.1. Then we introduce the neuron-level motivation and analysis in Section 4.2, and the Neuron-Fusion method in Section 4.3. 

\subsection{Locate-then-Merge Framework}
Model merging is an emerging research area that aims to combine the parameters of different models to achieve better performance. Early studies on model merging originate from model soups \cite{wortsman2022model}, which demonstrate that simply taking a weighted average of parameters from different models can improve overall performance: \[
\theta_{\text{merge}} = (1-\alpha) \theta_{\text{base}} + \alpha \theta_{\text{tuned}}
\]
This approach can also be interpreted from the perspective of Task Arithmetic \cite{ilharco2022editing}, where the merged model is expressed as:
\[
\theta_{\text{merge}} = \theta_{\text{base}} + \alpha \Delta
\]
with $\Delta = \theta_{\text{tuned}} - \theta_{\text{base}}$ representing the task vector that captures the parameter change. Building upon the task vector formulation, we propose the Locate-then-Merge framework for model merging:
\begin{equation}
    \theta_{\text{merge}} = \theta_{\text{base}} + F(\text{Sub}(\Delta))
\end{equation}
where $\text{Sub}(\cdot)$ is a function that locates a subset of parameters from $\Delta$, and $F(\cdot)$ is a function that transforms and merges the located parameters into the base model. The core intuition behind the Locate-then-Merge framework is inspired by the Lottery Ticket Hypothesis \cite{frankle2018lottery}, suggesting that a sparse subnetwork within a dense model can perform comparably to the full model. Similarly, we hypothesize that a carefully selected subset of parameter changes can effectively preserve the newly acquired capabilities.

The state-of-the-art model merging approaches can be viewed as special cases within the Locate-then-Merge framework. As summarized in Table 1, each method corresponds to a specific instantiation of the $\text{Sub}(\cdot)$ and $\text{F}(\cdot)$ functions, depending on how parameters are located and merged. The primary differences among these methods lie in the locating stage. TIES \cite{yadav2023ties} trims the task vectors based on their magnitudes and resolves sign conflicts to elect the final sign for each parameter. Breadcrumbs \cite{davari2024model} applies a sparse mask by removing parameters from the extreme tails of the absolute magnitude distribution. DARE \cite{yu2024language} randomly drops 99\% of the parameters. DELLA \cite{deep2024della} assigns higher dropout probabilities to parameters with lower magnitudes. At the merging stage, the main distinction is whether or not to rescale the selected parameters before combining them. 

\begin{table}[th]
\centering
\begin{tabular}{|c|c|c|}
\hline
\textbf{Method} & \textbf{Locate: Sub(x)} & \textbf{Merge: F(x)} \\
\hline
Task Ari & $x$ & $\alpha x$ \\
\hline
TIES & $\text{TRIMDrop}(x)$ & $\alpha x$ \\
\hline
Bread & $\text{TailDrop}(x)$ & $ \alpha x$ \\
\hline
DARE & $\text{RandomDrop}(x)$ & $\text{Rescale}(\alpha x)$\\
\hline
DELLA & $\text{MagDrop}(x) $ & $\text{Rescale}(\alpha x)$ \\
\hline
\end{tabular}
\caption{Mapping existing model merging methods into the Locate-then-Merge framework, categorized by locating (Sub) and merging (F) strategies.}
\end{table}

\subsection{Neuron-Level Motivation and Analysis}
\paragraph{Motivation for neuron-level fusion.} The existing methods primarily develop strategies to locate important parameters based on their individual magnitudes. However, they do not explicitly consider the structural roles of neurons. Recent studies \cite{dai2021knowledge,geva2022transformer,schwettmann2023multimodal,nikankin2024arithmetic} have shown that neurons serve as fundamental units that encode distinct capabilities. It is therefore highly plausible that certain neurons are disproportionately important during the fine-tuning process. Motivated by this insight, our work focuses on locating important neurons and designing neuron-level merging strategies to better preserve acquired capabilities while mitigating catastrophic forgetting.

\paragraph{Neuron change in FFN and attention layers.} We analyze the changes of FFN neurons and attention neurons between Llava-Next-Llama3 (tuned MLLM) \cite{liu2024improved} and Llama3 (base LLM) \cite{grattafiori2024llama}. In FFN layers, the $k$-th neuron corresponds to the $k$-th column in the down-projection matrix, as well as the $k$-th row in the up-projection and gate-projection matrices within the SwiGLU activation \cite{shazeer2020glu}. In attention layers, the $k$-th neuron represents the $k$-th column in the output matrix, as well as the $k$-th row in the query, key, and value matrices. For each neuron, we quantify its change by summing the absolute differences of its parameters across all dimensions:

\begin{equation}
    \text{C}(i) = \sum_{j} |\Delta_{i,j}|
\end{equation}
where $j$ indexes all dimensions (weights) associated with neuron $i$. $\Delta_{i,j}$ represents the parameter change on dimension $j$ in neuron $i$. The change of neurons in FFN and attention layers are shown in Figure 3 and 4. The x-axis is the neuron index, and the y-axis is the change score of the neurons. 

\begin{figure}[thb]
  \centering
  \includegraphics[width=0.98\columnwidth]{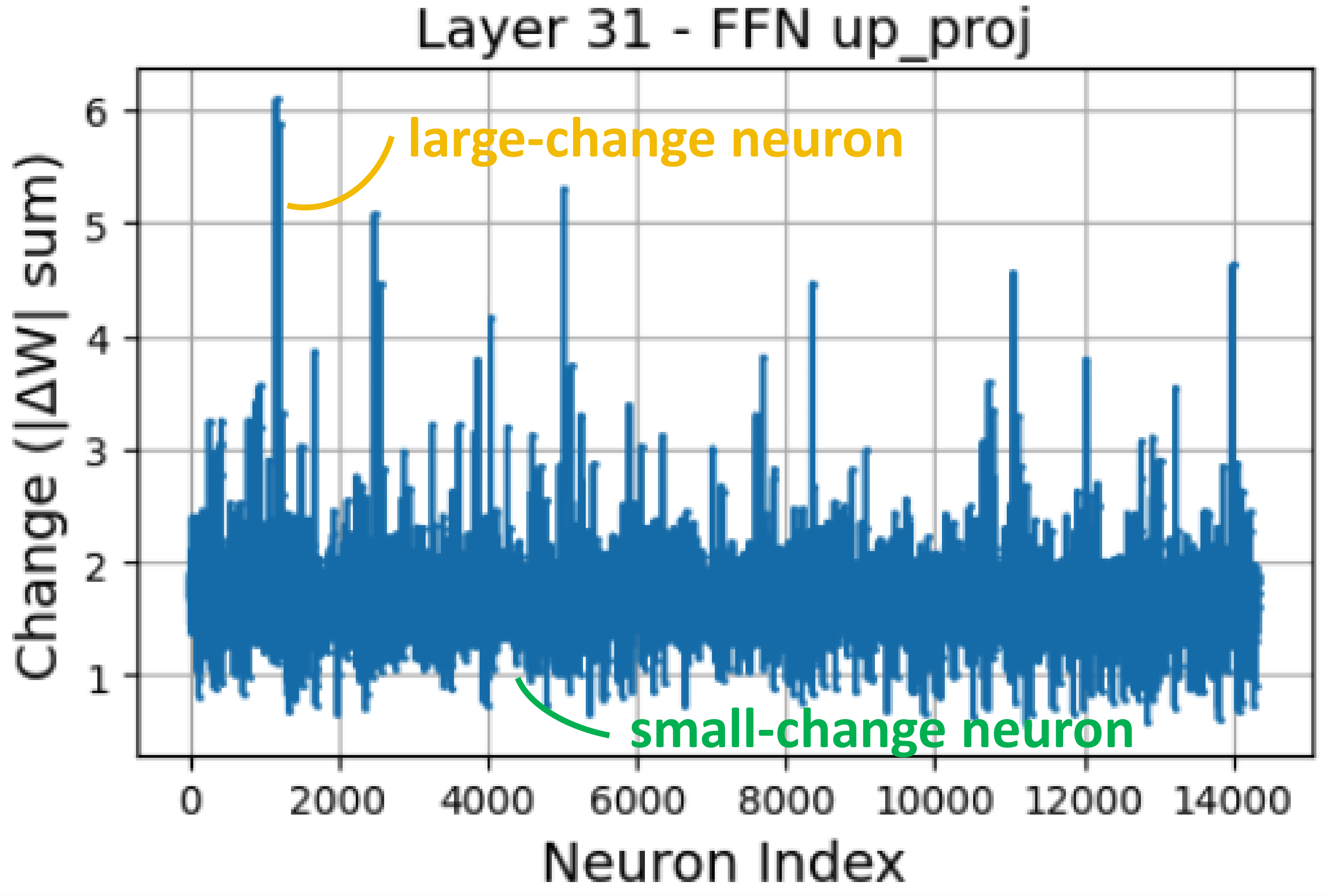}
  \caption{Change of neurons in FFN up matrix.}
\vspace{-10pt}
\end{figure}

\begin{figure}[thb]
  \centering
  \includegraphics[width=0.9\columnwidth]{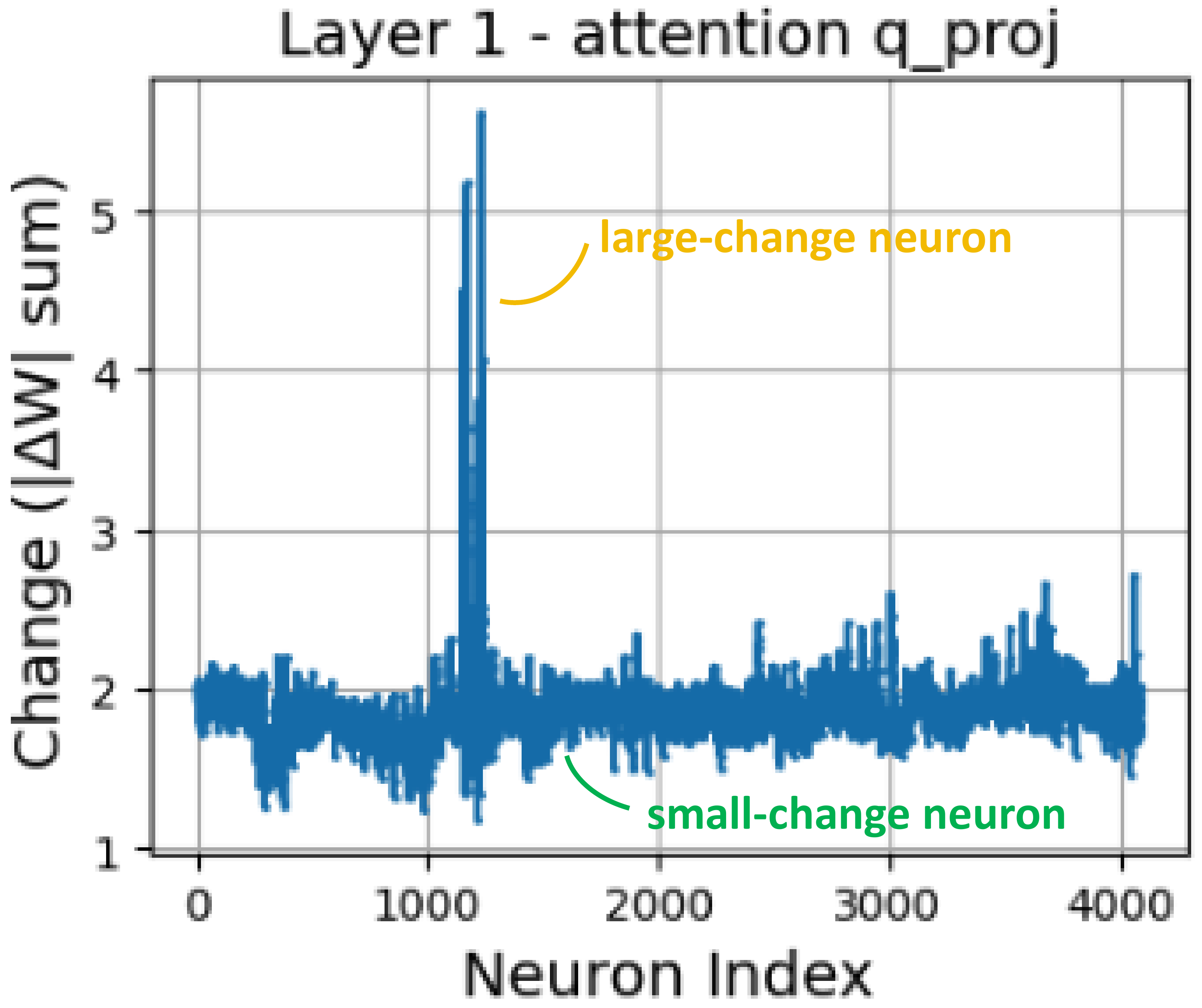}
  \caption{Change of neurons in attention query matrix.}
\vspace{-10pt}
\end{figure}

In Figure 3, we present the change of each neuron in the up-projection matrix in layer 31 as a representative example. Similarly, in Figure 4, we show the change of each neuron in the query matrix in layer 1. Similar trends are observed across other modules and layers. From the observation, we conclude that: \textbf{a small number of neurons exhibit significantly larger changes compared to the majority of neurons}.

\subsection{Neuron-Fusion Method for Mitigating Catastrophic Forgetting} 

To mitigate catastrophic forgetting, we propose Neuron-Fusion, a targeted neuron-level merging strategy. This method is based on a key hypothesis: neurons with large parameter changes during visual instruction tuning encode newly acquired visual capabilities, whereas widespread small changes may disrupt previously learned language abilities. Accordingly, Neuron-Fusion selectively preserves large-change neurons while suppressing small-change ones, as illustrated in Figure 1. Our approach consists of three steps:

(1) \textbf{Neuron-Locate}: We compute the change score $C(i)$ of each neuron by aggregating the absolute differences of its associated parameters (Eq. 2). Then we select the top $M\%$ neurons with the highest scores as candidates for preservation.

(2) \textbf{Neuron-Suppress}: To attenuate the effect of small-change neurons, we apply a parameter-level sparsification technique that retains only $K\%$ of parameters within each module. These retained parameters are scattered across neurons, which collectively reduces the influence of widespread parameter changes. This suppression step is agnostic to the specific sparsification method and can be implemented using TIES, Breadcrumbs, or other magnitude-based techniques.

(3) \textbf{Neuron-Restore}: To restore the contributions of the previously identified large-change neurons, we introduce two restoration strategies:

\textit{(a) Neuron-Replace}: Directly reinstates the parameters of the selected neurons from the tuned model:
\begin{equation}
    \Delta'_{i,j} = 
    \begin{cases}
    \Delta_{i,j}, & \text{if } |\Delta_{i,j}| \text{ is in } K\%  \\
    \Delta_{i,j}, & \text{if neuron } i \text{ is in } M\% \\
    0, & \text{otherwise}
    \end{cases}
\end{equation}

\textit{(b) Neuron-Rescale}: Adjusts the remaining parameters of each preserved neuron to match its original change score:
\begin{equation}
    \Delta'_{i,j} = 
    \begin{cases}
    0, & \text{if } |\Delta_{i,j}| \text{ is not in } K\%  \\
    \Delta_{i,j} \times \frac{C(i)}{C'(i)}, & \text{if neuron } i \text{ is in } M\% \\
    \Delta_{i,j}, & \text{otherwise}
    \end{cases}
\end{equation}
where $C'(i)$ is the new change score after suppression. The merged model's parameters are calculated by:
\begin{equation}
    \theta_{merge} = \theta_{base} + \Delta'
\end{equation}

In the Locate-then-Merge framework (Eq. 1), Neuron-Suppress corresponds to the locating function $\text{Sub}(\cdot)$, while Neuron-Restore (either Replace or Rescale) serves as the merging function $\text{F}(\cdot)$.

\begin{figure}[thb]
  \centering
  \includegraphics[width=0.98\columnwidth]{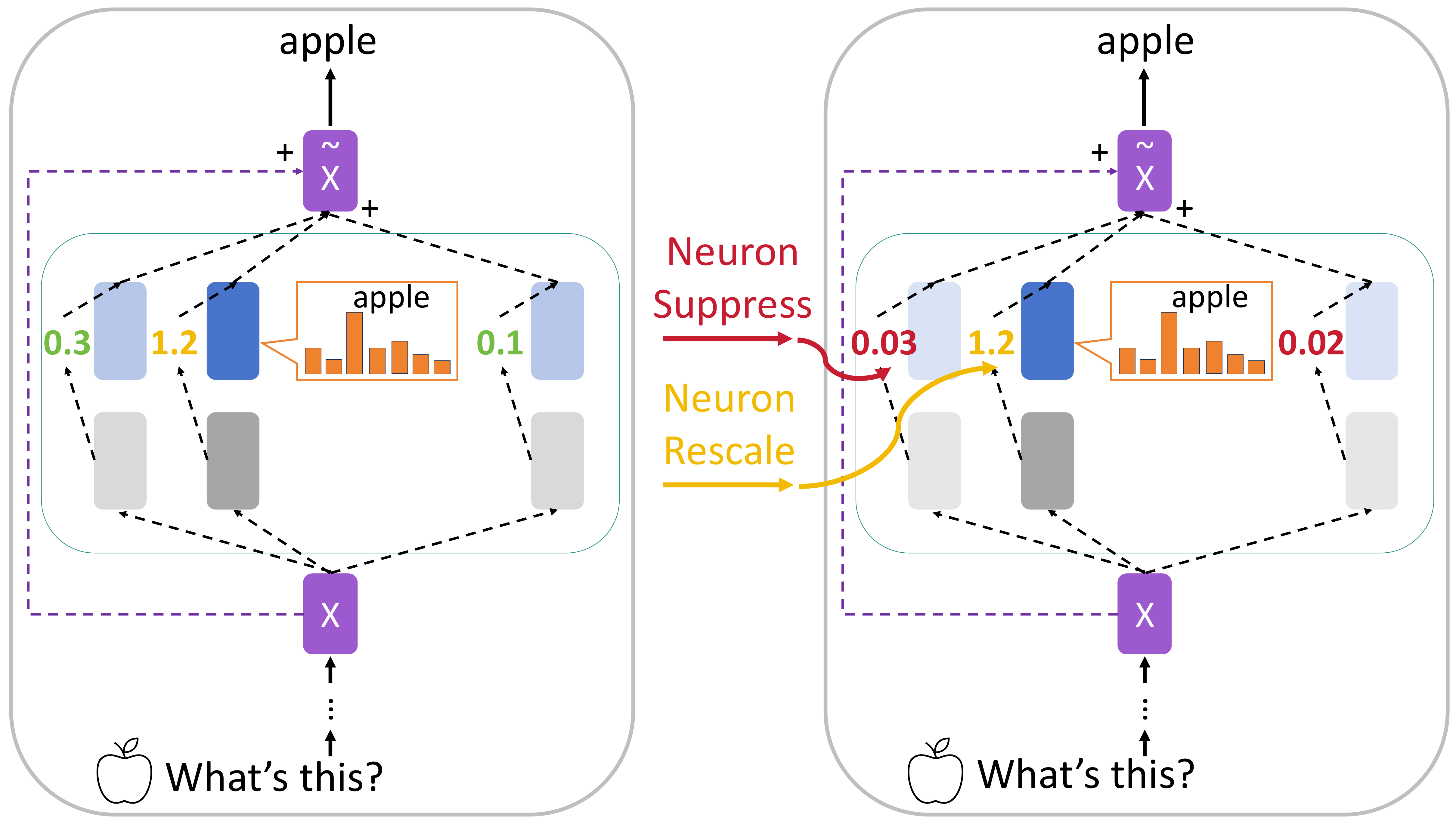}
  \caption{Change of coefficients after Neuron-Fusion.}
\vspace{-10pt}
\end{figure}

\paragraph{Key insight: restoring important neurons' coefficients via Neuron-Restore.} A key insight of Neuron-Restore comes from the lens of mechanistic interpretability, particularly drawing on the key-value memory view of FFN layers. Following \citet{geva2020transformer}, the two MLPs in a transformer FFN layer correspond to a set of subkeys (first MLP rows) and subvalues (second MLP columns). The FFN output is a weighted sum of subvalues, where the weights—called coefficients—are determined by the inner product between the input vector and each subkey. Furthermore, \citet{geva2022transformer} show that the distribution over final prediction is influenced directly by these subvalues. As illustrated in Figure 5 (left), the second neuron contributes most strongly to predicting ``apple''. Its large coefficient (1.2) amplifies the corresponding subvalue, shifting the output distribution toward the correct label.

Now consider how Neuron-Fusion alters this mechanism, shown in Figure 5 (right). Let $x \in \mathbb{R}^D$ be the FFN input and $\Delta_i = [\Delta_{i,1}, \ldots, \Delta_{i,D}]$ be the parameter changes of the $i$-th subkey. The original coefficient $c_i$ is computed as:
\[
c_i = \sum_{j=1}^D x_j \Delta_{i,j}
\]
When we apply the Neuron-Suppress stage, only $K\%$ dimensions of $\Delta_i$ are retained, reducing the coefficient to $\tilde{c}_i$:
\[
\tilde{c}_i = \sum_{j \in S} x_j \Delta_{i,j}, \quad S \subset \{1,\ldots,D\}
\]
This approximation weakens the contribution of important neurons and may reduce the probability of correct answers. To counteract this, the Neuron-Replace step can replace $\Delta_{i,j}$ into the important neurons and restore $c_i$. The Neuron-Rescale step can enlarge the surviving parameters in each important neuron to restore the total change score, where the new coefficient $\hat{c}_i$ becomes:
\[
\hat{c}_i = \frac{\sum_{j=1}^D |\Delta_{i,j}|}{\sum_{j \in S} |\Delta_{i,j}|} \cdot \tilde{c}_i
\]
If the input vector $x$ has roughly uniform values, this rescaling approximately restores $\hat{c}_i \approx c_i$, thus preserving the neuron’s influence.

\section{Experiments}
\subsection{Experimental Settings}
\paragraph{Language ability evaluation.} We choose seven widely used benchmark datasets—RACE \cite{lai2017race}, CommonsenseQA \cite{talmor2018commonsenseqa}, PIQA \cite{bisk2020piqa}, OpenbookQA \cite{mihaylov2018can}, GSM8K \cite{cobbe2021training}, ARC-Easy, and ARC-Challenge \cite{clark2018think}—to evaluate the general language ability of the model. 
These datasets cover a broad range of language understanding tasks, including reading comprehension, commonsense reasoning, mathematical problem solving, and general science question answering. Together, they provide a comprehensive assessment of a model's language capabilities. We use 8-shot chain-of-thought for GSM8K, and zero-shot for other datasets. We use the lm-evaluation-harness \cite{eval-harness} library to calculate the exact-match accuracy for all the datasets. 

\paragraph{Visual ability evaluation.} We choose six common benchmarks—MME \cite{fu2023mme}, MMMU \cite{yue2024mmmu}, ScienceQA \cite{lu2022learn}, GQA \cite{hudson2019gqa}, MMBench-CN, and MMBench-EN \cite{liu2024mmbench}—to evaluate the visual ability of the model. These datasets cover a wide range of vision understanding skills, including fine-grained perception, subject-specific reasoning across science, math, and humanities, vision science question answering, and visual reasoning. We use the lmms-eval \cite{lmms_eval2024} library to calculate the exact-match accuracy for these datasets.

\paragraph{Models.} Model merging methods require access to the parameters of LLMs and MLLMs. So we choose two powerful open-source LLMs—Mistral-7B \cite{jiang2024identifying} and Llama3-8B \cite{grattafiori2024llama}—as our base models. The corresponding MLLMs are obtained by performing visual instruction tuning \cite{liu2024improved} on vision-language datasets. We use MergeKit \cite{goddard2024arcee} library to merge the models.

\begin{table}[ht]
\centering
\begin{tabular}{lccc}
\toprule
\textbf{Method} & \textbf{LA (\%)} & \textbf{VA (\%)} & \textbf{OA (\%)} \\
\midrule
MLLM & 57.14 & 64.76 & 60.95 \\
LLM & 62.91 & 48.63 & 55.77 \\
\midrule
Task Ari & 61.39 & 63.90 & 62.65 \\
TIES & 62.43 & 62.19 & 62.31 \\
Breadcrumbs & 60.20 & 64.60 & 62.40 \\
DARE & 56.91 & 64.93 & 60.92\\
DELLA & 55.60 & 64.12 & 59.86 \\
\midrule
Neu-P-TaskA & 61.89 & 63.90 & \textbf{62.90} \\
Neu-P-TIES & 61.80 & 63.27 & 62.54 \\
Neu-S-TIES & 61.2 & 63.5 & 62.35 \\
Neu-P-Bread & 62.00 & 63.53 & 62.77 \\
Neu-S-Bread & 61.74 & 63.40 & 62.57 \\
\bottomrule
\end{tabular}
\caption{Comparison of Neuron-Fusion and other model merging methods on Llama3. We report LA (language ability), VA (visual ability), and OA (overall ability).}
\end{table}

\begin{table}[ht]
\centering
\begin{tabular}{lccc}
\toprule
\textbf{Method} & \textbf{LA (\%)} & \textbf{VA (\%)} & \textbf{OA (\%)} \\
\midrule
MLLM & 55.40 & 63.07 & 59.23 \\
LLM & 56.89 & 32.53 & 44.71 \\
\midrule
Task Ari & 56.86 & 63.16 & 60.01 \\
TIES & 57.60 & 62.57 & 60.08 \\
Breadcrumbs & 57.46 & 62.53 & 59.99 \\
DARE & 55.74 & 63.13 & 59.43 \\
DELLA & 53.63 & 62.27 & 57.95 \\
\midrule
Neu-P-TaskA & 57.80 & 62.73 & \textbf{60.27} \\
Neu-P-TIES & 57.34 & 62.43 & 59.89 \\
Neu-S-TIES & 56.85 & 62.67 & 59.76 \\
Neu-P-Bread & 57.09 & 63.07 & 60.08 \\
Neu-S-Bread & 57.31 & 61.86 & 59.59 \\
\bottomrule
\end{tabular}
\caption{Comparison of Neuron-Fusion and other model merging methods on Mistral. We report LA (language ability), VA (visual ability), and OA (overall ability).}
\end{table}

\subsection{Results of Neuron-Fusion Method}
\paragraph{Comparative Performance.}

We evaluate Neuron-Fusion against several state-of-the-art model merging methods discussed in Section 4.1, including Task Arithmetic, TIES, Breadcrumbs, DARE, and DELLA. For Neuron-Fusion, we use Task Arithmetic, TIES, and Breadcrumbs as the Neuron-Suppress strategy and integrate them with either Neuron-Replace or Neuron-Rescale during the Neuron-Restore phase. These configurations are denoted as Neu-P-* and Neu-S-*, respectively.

Table 2 and Table 3 summarize the results on Llama3 and Mistral models, and the detailed results are shown in Appendix A. Neuron-Fusion consistently achieves the best overall ability (OA), demonstrating its effectiveness in simultaneously preserving language ability (LA) and visual ability (VA). Notably, Neu-P-TaskA and Neu-P-Bread achieve the highest scores on both models, showing that restoring high-impact neurons yields the most balanced performance. In contrast, DARE and DELLA exhibit inferior performance, with LA scores even lower than the base LLMs. This suggests that indiscriminate parameter dropout or excessive rescaling can severely disrupt learned representations.

\begin{figure}[htb]
  \centering
  \includegraphics[width=0.95\columnwidth]{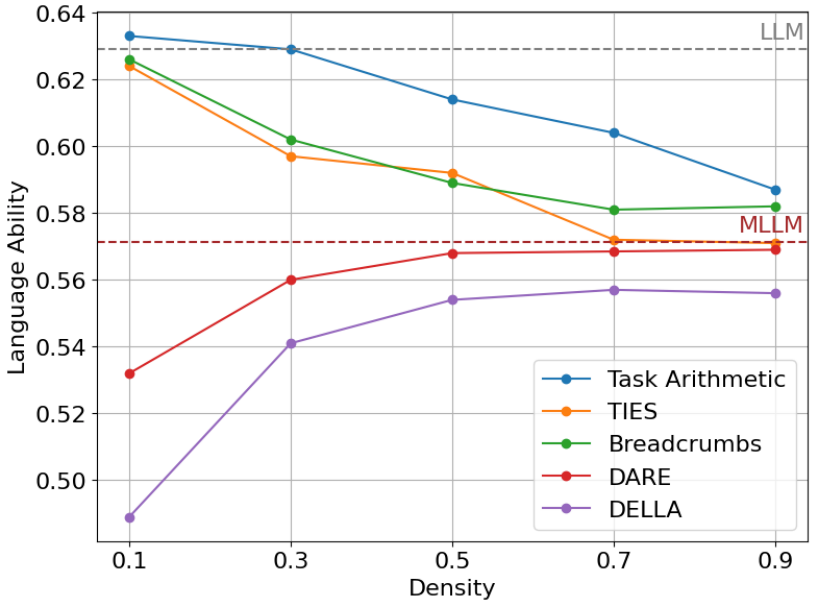}
  \caption{Language ability under different density K.}
\vspace{-10pt}
\end{figure}

\begin{figure}[htb]
  \centering
  \includegraphics[width=0.95\columnwidth]{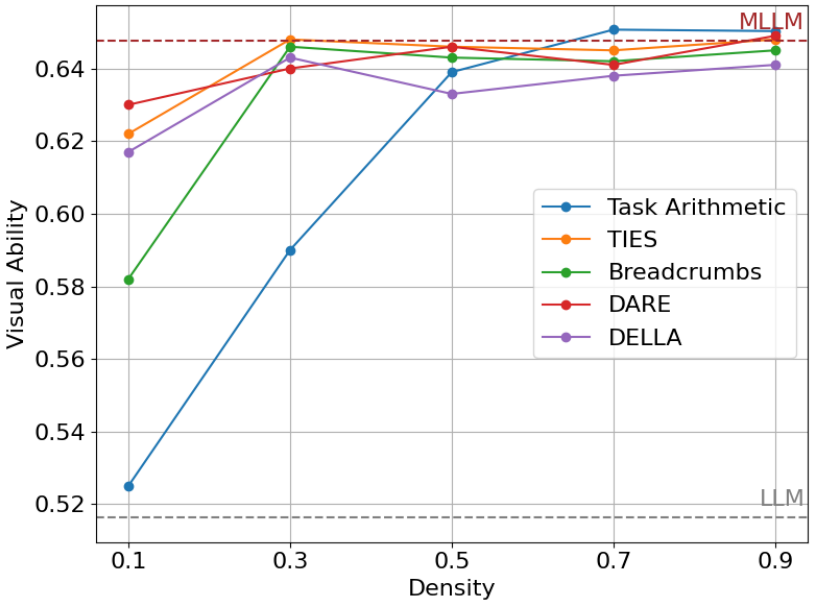}
  \caption{Visual ability under different density K.}
\vspace{-10pt}
\end{figure}

\paragraph{Ablation on Neuron-Suppression density $K\%$.} We conduct an ablation study to analyze the impact of the suppression density hyperparameter $K\%$, which controls the proportion of parameters retained during the Neuron-Suppress stage. As shown in Figure 6 and 7, we evaluate multiple merging strategies across different $K\%$ on Llama3. The curves of Mistral are shown in Appendix B, which have similar trends. On language ability, DARE and DELLA consistently perform worse than the MLLM across all densities, indicating they fail to recover even the degraded language skills. In contrast, Task Arithmetic, TIES, and Breadcrumbs show monotonically decreasing performance. On visual ability, all methods reach their lowest performance at K=0.1, then improve and decline again—showing a inflection point. Identifying this inflection point is crucial for balancing language retention and visual adaptation. 

\begin{figure}[htb]
  \centering
  \includegraphics[width=0.95\columnwidth]{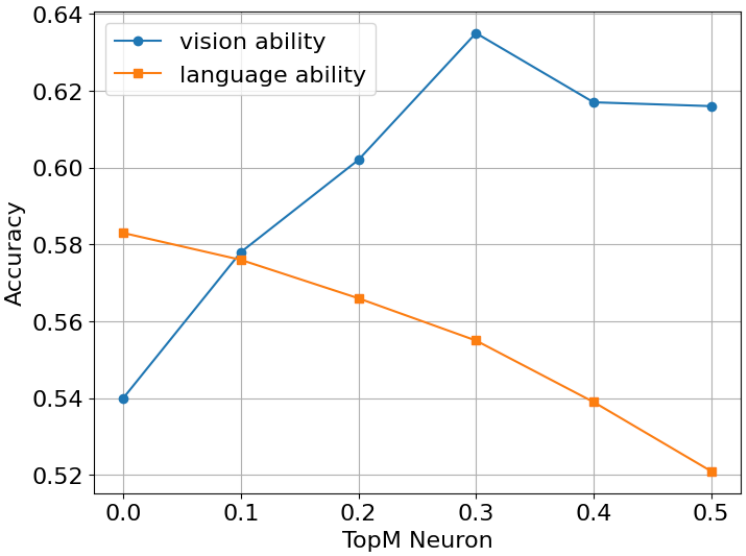}
  \caption{Results when restoring TopM (\%) neurons.}
\vspace{-10pt}
\end{figure}

\paragraph{Ablation on Neuron-Restore ratio $M\%$.} We further study the impact of the neuron restoration ratio $M\%$, which controls the fraction of neurons whose changes are restored. Using TIES with K=0.2 and Neuron-Rescale, we vary $M$ and plot the results in Figure 8. When $M = 0.0$, the vision ability is low due to the suppression of all neuron-level changes. As $M$ increases from 0.0 to 0.5, the language ability decreases gradually, while the vision ability improves sharply up to $M = 0.3$ and then begins to decline. This trend indicates that a moderate value of $M$ achieves a good balance between retaining visual capabilities and mitigating catastrophic forgetting. These results align with our hypothesis that large-change neurons are more critical for storing visual capabilities. Restoring the large-change neurons enhances vision ability, but extending restoring to more neurons introduces diminishing returns and may even harm performance.

\paragraph{Generation analysis.} To better understand how Neuron-Fusion improves the accuracy in visual datasets, we analyze the generations at two stages of the Neuron-Fusion process. The most significant improvement occurs on the ScienceQA dataset. The accuracy after hard-merge is 0\%. After Neuron-Suppress with $K = 0.2$, the accuracy increases to 7\%. After Neuron-Restore with $M = 0.3$, the accuracy improves substantially to 52.6\%. To investigate this improvement, we examine the transitions from false to correct generations between the stages of 0\% to 7\%, and 7\% to 52.6\%. We identify two major types of failure cases, as illustrated in Table 4. The first two examples fall under the "Not-Known" category, where the model refuses to answer. The last two examples are categorized as "Context-Hallucination," where the model produces content not grounded in the input. For instance, in the final example of Table 4, the false answer ``flamboyant cuttlefish'' does not correspond to any of the choices in the question.

\begin{table}[ht]
\centering
\small
\begin{tabular}{p{0.4\linewidth} p{0.25\linewidth} p{0.15\linewidth}}
\toprule
\textbf{Question \& Choices} & \textbf{False answer} & \textbf{Correct answer } \\
\midrule
Which continent is highlighted? A. Africa; B. North America; C. South America; D. Asia & I cannot directly see the image or the answer choices & D \\
\midrule
Which closing is correct for a letter? A. see you soon, Rose; B. See you soon, Rose & I cannot directly answer from the choices & B \\
\midrule
Which figure of speech is used in this text?
Sing, O goddess, the anger of Achilles son of Peleus A. chiasmus; B. apostrophe & B. Some of the grass on the ground is burning. & B \\
\midrule
Which animal's skin is better adapted as a warning sign? A. lichen katydid; B. opalescent nudibranch & B. flamboyant cuttlefish & B \\
\bottomrule
\end{tabular}
\caption{Examples of false to correct generations on the ScienceQA dataset after Neuron-Fusion: "Not-Known" (first 2) and "Context-Hallucination" (last 2) problems.}
\end{table}

We find that the Neuron-Suppress stage primarily addresses the "Not-Known" problem: 56.7\% of the corrected cases at this stage fall into this category. In contrast, the Neuron-Restore stage is particularly effective at resolving "Context-Hallucination" errors, correcting 97.2\% of such cases in this stage. These findings highlight that Neuron-Fusion effectively mitigates two critical issues in multimodal generation: uncertainty in answering and hallucination beyond provided context, which demonstrates that our method not only recovers quantitative performance but also qualitatively enhances output consistency and trustworthiness.

\section{Conclusion}
We propose Locate-then-Merge, a framework for mitigating catastrophic forgetting in MLLMs. Based on this framework, we develop Neuron-Fusion, a neuron-level parameter fusion method that selectively preserves neurons with large parameter shifts while suppressing harmful widespread changes. Through extensive experiments across language and vision benchmarks, Neuron-Fusion consistently outperforms existing model merging techniques, achieving better retention of both language and visual capabilities. Furthermore, generation analysis reveals that our method effectively reduces common failure modes such as Not-Known and Context-Hallucination, leading to more reliable and controllable model outputs. These results demonstrate the potential of neuron-level fusion strategies for advancing MLLMs' abilities.

\clearpage

\section*{Limitations}
Our work focuses on the vision modality and the visual instruction tuning paradigm for MLLMs. While the proposed Neuron-Fusion method demonstrates strong performance on vision-language benchmarks, we have not investigated whether the same approach can be effectively extended to other modalities such as audio or video. Also, we do not examine its applicability to alternative vision-language model architectures such as CLIP. Additionally, our method is specifically developed for decoder-only LLMs, which currently represent the dominant architecture in high-performing language and multimodal models. Future work is needed to evaluate the generalizability of our approach across diverse multimodal frameworks and architectures.

The Locate-then-Merge framework and the Neuron-Fusion method are specifically designed for scenarios involving parameter merging between a base model and its fine-tuned counterpart, assuming both models share the same architecture and differ only in learned weights. As such, our method may not be directly applicable to settings involving architectural discrepancies or to merging independently trained models with differing objectives or tasks. Extending the framework to accommodate more diverse model merging scenarios remains an important avenue for future exploration.

\bibliography{custom}

\clearpage
\appendix
\section{Detailed Results on All Datasets}
\subsection{Catastrophic Forgetting in MLLMs}
\begin{figure}[htbp]
    \centering
    \begin{minipage}[t]{0.23\textwidth}
        \centering
        \includegraphics[width=\textwidth]{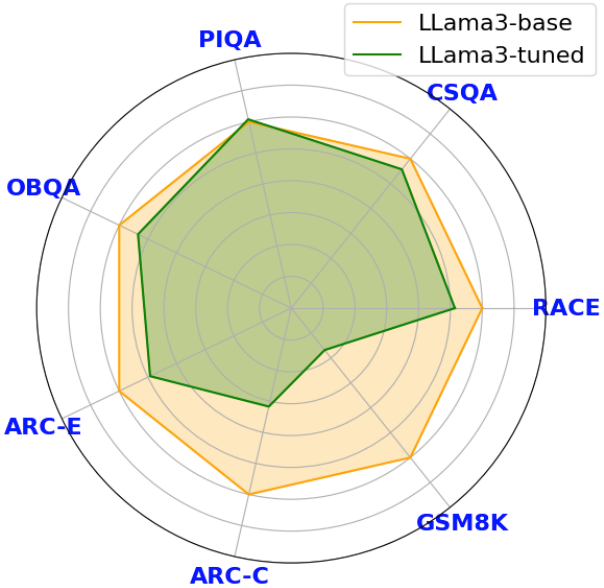}
    \end{minipage}
    \hfill
    \begin{minipage}[t]{0.23\textwidth}
        \centering
        \includegraphics[width=\textwidth]{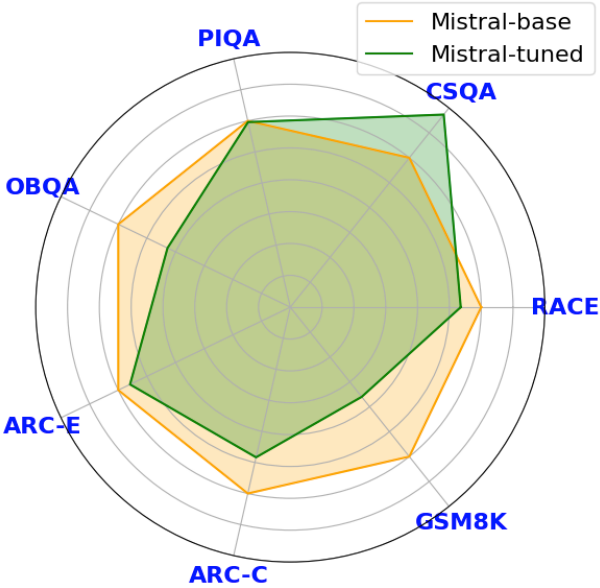}
    \end{minipage}
    \caption{Accuracy of Llama3 (left) and Mistral (right) on language datasets after visual instruction tuning.}
\end{figure}

We present the accuracy of Llama3 and Mistral on the language datasets before and after visual instruction tuning in Figure 9. Except for Mistral's result on CommonsenseQA, the accuracy consistently decreases across most datasets, with particularly significant drops observed on GSM8K and ARC-Challenge. These results confirm the presence of catastrophic forgetting in the language capabilities after visual instruction tuning.

\begin{figure}[htbp]
    \centering
    \begin{minipage}[t]{0.23\textwidth}
        \centering
        \includegraphics[width=\textwidth]{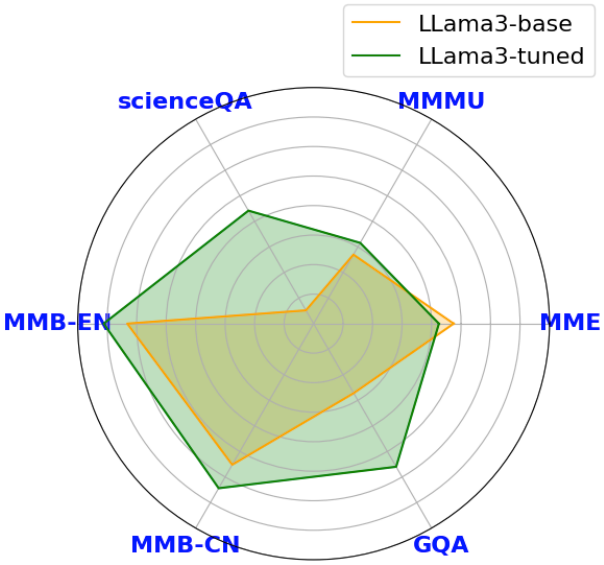}
    \end{minipage}
    \hfill
    \begin{minipage}[t]{0.23\textwidth}
        \centering
        \includegraphics[width=\textwidth]{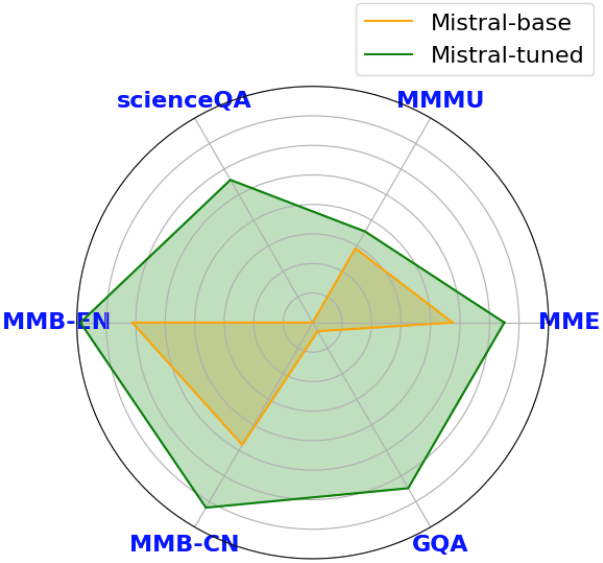}
    \end{minipage}
    \caption{Accuracy of Llama3 (left) and Mistral (right) on vision datasets after hard-merge into tuned MLLMs.}
\end{figure}

As introduced in Section 3.2, a straightforward approach for recovering language ability is to directly replace the tuned MLLM's parameters with the base LLM. In Figure 10, we present the accuracy on visual datasets after hard-merging the original parameters of Llama3 and Mistral into their corresponding tuned MLLMs. Across both models and datasets, the accuracy drops significantly. Therefore, it is necessary to design a more effective method that can restore the language capabilities while preserving the visual capabilities.

\subsection{Results on All Datasets}

The detailed results on 6 visual datasets and 7 language datasets in Llama3 and Mistral are shown in Table 5 and Table 6, respectively.

\begin{table*}[htbp]
\centering
\begin{adjustbox}{max width=\textwidth}
\begin{tabular}{l|cccccc|ccccccccc}
\toprule
Method & MME & MMMU & SciQA & MMB-E & MMB-C & GQA & RACE & CSQA & PIQA & OBQA & ARC-E & ARC-C & GSM8K \\
\midrule
MLLM & 71.2 & 41.6 & 54.2 & 81.2 & 74.4 & 66.0 & 42.2 & 73.0 & 79.2 & 32.8 & 74.8 & 42.8 & 55.2 \\
LLM & 64.0 & 37.0 & 15.2 & 73.2 & 65.2 & 37.2 & 44.6 & 75.0 & 78.8 & 34.2 & 80.2 & 52.0 & 75.6 \\
\midrule
Task Arithm & 71.8 & 39.6 & 55.8 & 81.8 & 74.6 & 59.8 & 43.8 & 76.8 & 79.8 & 35.6 & 79.0 & 48.6 & 66.2 \\
TIES & 70.4 & 41.0 & 53.2 & 78.8 & 73.4 & 56.4 & 44.8 & 78.0 & 79.2 & 36.2 & 79.0 & 49.0 & 70.8 \\
Breadcrumbs & 71.0 & 40.8 & 55.6 & 82.0 & 74.0 & 64.2 & 43.8 & 76.6 & 79.6 & 34.8 & 75.8 & 46.6 & 64.2 \\
DARE & 71.8 & 41.6 & 54.4 & 81.0 & 74.4 & 66.4 & 42.4 & 72.8 & 79.2 & 32.8 & 74.8 & 42.2 & 54.2 \\
DELLA & 70.4 & 39.8 & 54.8 & 79.6 & 74.4 & 65.6 & 42.0 & 71.2 & 77.8 & 32.8 & 72.6 & 41.2 & 51.6 \\
\midrule
Neu-P-TaskA & 71.0 & 41.4 & 57.6 & 81.0 & 74.2 & 58.2  & 44.0 & 77.8 & 79.8 & 35.2 & 79.4 & 49.4 & 67.6  \\
Neu-P-TIES   & 70.4 & 40.6 & 54.4 & 80.4 & 75.2 & 58.6 & 43.6 & 77.8 & 78.8 & 36.0 & 78.2 & 48.6 & 69.6 \\
Neu-S-TIES   & 70.6 & 40.2 & 54.4 & 80.4 & 75.4 & 60.0 & 43.8 & 78.0 & 79.4 & 35.0 & 77.4 & 47.0 & 67.8 \\
Neu-P-Bread  & 70.0 & 40.8 & 56.2 & 80.0 & 75.0 & 59.2 & 44.0 & 78.2 & 78.6 & 36.0 & 78.0 & 48.4 & 70.8 \\
Neu-S-Bread & 70.0 & 40.0 & 57.0 & 80.2 & 75.0 & 58.2 & 44.4 & 77.0 & 78.6 & 35.0 & 78.8 & 47.8 & 70.6 & \\
\bottomrule
\end{tabular}
\end{adjustbox}
\caption{Results of different methods on 6 visual datasets and 7 language datasets in Llama3-7B.}
\end{table*}

\begin{table*}[htbp]
\centering
\begin{adjustbox}{max width=\textwidth}
\begin{tabular}{l|cccccc|ccccccc}
\toprule
Model & MME & MMMU & SQA & MMB-E & MMB-C & GQA & RACE & CSQA & PIQA & OBQA & ARC-E & ARC-C & GSM8K \\
\midrule
MLLM & 71.0 & 35.6 & 55.8 & 78.8 & 72.4 & 64.8 & 43.0 & 73.6 & 81.2 & 31.6 & 76.0 & 48.4 & 34.0 \\
LLM  & 53.8 & 29.0 & 0.0  & 61.2 & 47.8 & 3.4  & 44.8 & 66.4 & 81.4 & 35.4 & 78.0 & 52.2 & 40.0 \\
\midrule
Task Arithm  & 71.4 & 36.0 & 56.8 & 79.4 & 72.0 & 63.4 & 43.6 & 74.2 & 82.2 & 32.8 & 77.0 & 49.4 & 38.8 \\
TIES         & 72.8 & 37.0 & 53.8 & 78.0 & 73.0 & 60.8 & 44.0 & 74.0 & 83.0 & 32.6 & 78.4 & 51.0 & 40.2 \\
Breadcrumbs  & 71.6 & 34.8 & 56.0 & 78.4 & 72.2 & 62.2 & 44.2 & 74.2 & 82.6 & 33.0 & 78.2 & 50.8 & 39.2 \\
DARE         & 71.4 & 35.8 & 56.0 & 79.0 & 72.0 & 64.6 & 42.8 & 73.6 & 81.6 & 31.8 & 75.8 & 48.2 & 36.4 \\
DELLA        & 70.2 & 33.2 & 55.8 & 78.6 & 70.6 & 65.2 & 41.8 & 72.6 & 80.6 & 30.2 & 74.2 & 46.4 & 29.6 \\
\midrule
Neu-P-TaskA & 72.8 & 36.6 & 53.6 & 79.2 & 73.2 & 61.0 & 43.8 & 74.2 & 82.8 & 33.2 & 78.6 & 51.8 & 40.2 \\
Neu-P-TIES   & 72.6 & 36.0 & 54.2 & 78.0 & 72.8 & 61.0 & 44.0 & 73.4 & 83.0 & 32.6 & 78.0 & 50.8 & 39.6 \\
Neu-S-TIES   & 72.8 & 36.2 & 54.2 & 78.0 & 73.0 & 62.2 & 43.4 & 72.4 & 83.0 & 31.4 & 77.6 & 50.4 & 39.8 \\
Neu-P-Bread  & 72.2 & 35.8 & 57.0 & 79.0 & 72.0 & 62.4 & 44.2 & 73.8 & 82.4 & 33.0 & 77.0 & 49.6 & 39.6 \\
Neu-S-Bread  & 72.2 & 35.6 & 51.4 & 79.0 & 72.0 & 61.0 & 43.6 & 73.4 & 83.2 & 33.0 & 78.2 & 50.8 & 39.0 \\
\bottomrule
\end{tabular}
\end{adjustbox}
\caption{Results of different methods on 6 visual datasets and 7 language datasets in Mistral-7B.}
\end{table*}

\section{Different Density in Mistral}
The language and visual capabilities of Task Arithmetic, TIES, Breadcrumbs, DARE, and DELLA under different density $K\%$ in Mistral are illustrated in Figures 11 and 12. These results exhibit trends similar to those observed with Llama3. A difference is that the Mistral-based methods—Task Arithmetic, TIES, and Breadcrumbs—achieve higher accuracy than the standalone LLM when the density is low. This is attributed to the MLLM achieving better performance on the CommonsenseQA dataset compared to the LLM (see Table 6). By integrating the MLLM with the LLM, the combined model benefits from the performance gains on this dataset.
\begin{figure}[htb]
  \centering
  \includegraphics[width=0.95\columnwidth]{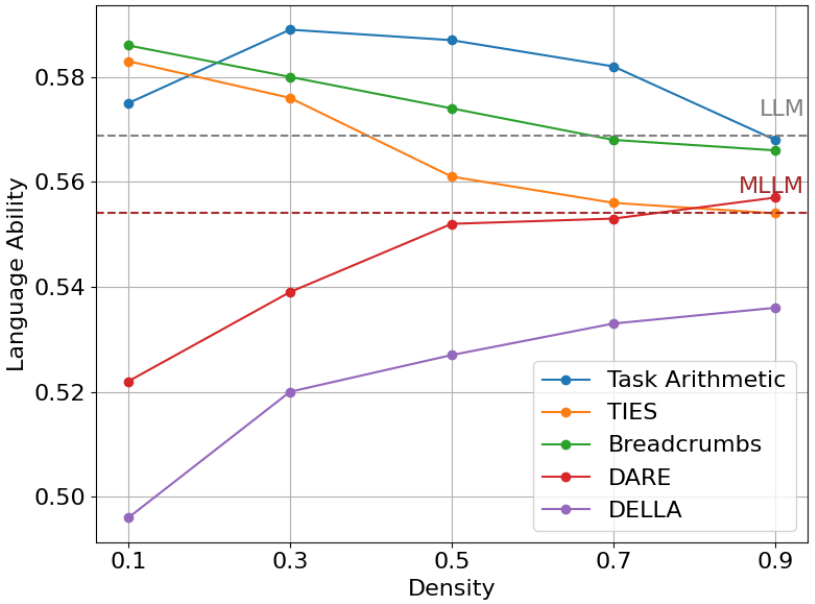}
  \caption{Language ability under different density K.}
\vspace{-10pt}
\end{figure}

\begin{figure}[htb]
  \centering
  \includegraphics[width=0.95\columnwidth]{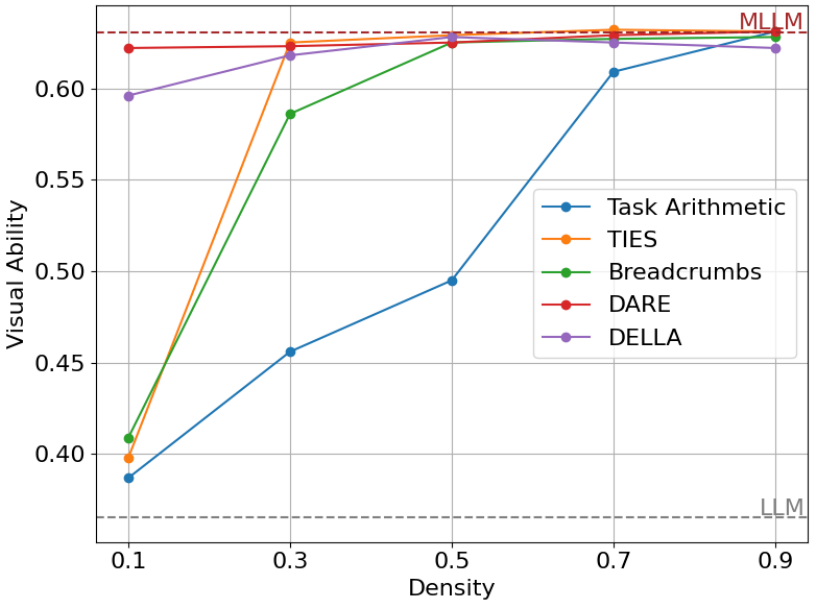}
  \caption{Visual ability under different density K.}
\vspace{-10pt}
\end{figure}

\end{document}